\PassOptionsToPackage{utf8}{inputenc}

\documentclass{bioinfo}

\usepackage{subfigure}
\usepackage{subfig} 

\usepackage{multirow}
\usepackage{xcolor}
\copyrightyear{2022} \pubyear{2022}

\access{Advance Access Publication Date: Day Month Year}
\appnotes{Original Paper}

\begin{document}
\firstpage{1}

\subtitle{Data and text mining}

\title[short Title]{Hierarchical Reinforcement Learning for Automatic Disease Diagnosis}
\author[Sample \textit{et~al}.]{Cheng Zhong\,$^{\text{\sfb 1},\&}$ , Kangenbei Liao\,$^{\text{\sfb 1},\&}$, Wei Chen\,$^{\text{\sfb 1}}$, Qianlong Liu\,$^{\text{\sfb 2}}$,  Baolin Peng\,$^{\text{\sfb 3}}$, Xuanjing Huang\,$^{\text{\sfb 5}}$,
Jiajie Peng\,$^{\text{\sfb 4}*}$ and Zhongyu Wei\,$^{\text{\sfb 1,4,}*}$,}
\address{$^{\text{\sf 1}}$School of Data Science, Fudan University, China \\
$^{\text{\sf 2}}$Alibaba Group, China \\
$^{\text{\sf 3}}$Microsoft Research, USA \\
$^{\text{\sf 4}}$Research Institute of Intelligent Complex Symtems, Fudan University, China\\
$^{\text{\sf 5}}$School of Computer Science, Fudan University, China \\}

\corresp{$^\ast$To whom correspondence should be addressed.\\
$^\&$These authors contributed equally to this work and should be considered co-first authors.}

\history{Received on 21 March 2022; revised on 16 May 2022; accepted on 29 June 2022}

\editor{Associate Editor: Jonathan Wren}

\abstract{\textbf{Motivation:} Disease diagnosis oriented dialogue system models the interactive consultation procedure as Markov Decision Process and reinforcement learning algorithms are used to solve the problem. Existing approaches usually employ a flat policy structure that treat all symptoms and diseases equally for action making. This strategy works well in the simple scenario when the action space is small, however, its efficiency will be challenged in the real environment. Inspired by the offline consultation process, we propose to integrate a hierarchical policy structure of two levels into the dialogue system for policy learning. The high-level policy consists of a master model that is responsible for triggering a low-level model, the low-level policy consists of several symptom checkers and a disease classifier. The proposed policy structure is capable to deal with diagnosis problem including large number of diseases and symptoms. \\
\textbf{Results:} Experimental results on three real-world datasets and a synthetic dataset demonstrate that our hierarchical framework achieves higher accuracy and symptom recall in disease diagnosis compared with existing systems. We construct a benchmark including datasets and implementation of existing algorithms to encourage follow-up researches.\\
\textbf{Availability:} The code and data is available from  https://github.com/FudanDISC/DISCOpen-MedBox-DialoDiagnosis\\
\textbf{Contact:} \href{name@bio.com}{21210980124@m.fudan.edu.cn}\\
\textbf{Supplementary information:} Supplementary data are available at \textit{Bioinformatics}
online.}

\maketitle

\section{Introduction}

With the development of electronic medical records (EMRs), researchers have explored different machine learning approaches for automatic diagnosis~\citep{shivade2013review, richens2022artificial}. Although impressive results have been reported for the identification of various diseases, e.g. heart failure with preserved ejection fraction (HFpEF)~\citep{jonnalagadda2017text} and autism spectrum disorders (ASDs)~\citep{doshi2013comorbidity}, they rely on well-established EMRs which are labor-intensive to build. Moreover, the automatic diagnosis of a certain disease requires EMRs of that type for model training, and is difficult to be extended to other types of diseases.

To relieve the pressure of constructing EMRs, researchers~\citep{wei2018task,xu2019end} introduce the task-oriented dialogue system to request symptoms automatically from patients for disease diagnosis. Since the disease consultation is an interactive procedure with multiple time steps, they formulate the task as Markov Decision Processes (MDPs) and employ reinforcement learning (RL) based methods for the policy learning. In each round of interaction, the agent either chooses a symptom to request or makes the diagnosis via selecting an action from the joint action space of symptoms and diseases. Correct symptom query and disease diagnosis will be rewarded, and the policy is learned by maximizing the expected cumulative rewards. 

After that, reinforcement learning becomes the first choice for researchers in this field ~\citep{tang2016inquire,kao2018context,peng2018refuel, coronato2020reinforcement, yu2021reinforcement}. \citep{kao2018context} presented a context-aware hierarchical reinforcement method, using policy gradients to make decisions based on the patient’s personal information and explicit symptoms. \citep{peng2018refuel} proposed reward shaping and feature rebuilding techniques to help agents effectively learn a better strategy and \citep{2019Effective} introduced a new multiple action policy representation to help agents suggest medical tests to facilitate disease diagnosis. Also, many researchers tried to combine the RL-based and non-RL approaches. ~\citep{xu2019end} introduced the knowledge graph into their dialogue system, ~\citep{xia2020generative} applied GAN as a policy network to capture the relations between different symptoms, and ~\citep{lin2020towards} proposed DSMAD method which inspired by the introspective decision-making process of human to make the diagnosis process more reliable. Recently, ~\citep{hou2021imperfect} proposed a multi-level reward modeling approach and ~\citep{teixeira2021interplay} proposed an approach to automating the generation of a dialogue manager to achieve the predictability and reliability.
However, existing policies are designed with flat and monolithic structures (such as MLPs), which are not salable to deal with scenarios including a large number of diseases and symptoms.

\begin{figure*}[t]
\begin{center}
\includegraphics[width=\linewidth]{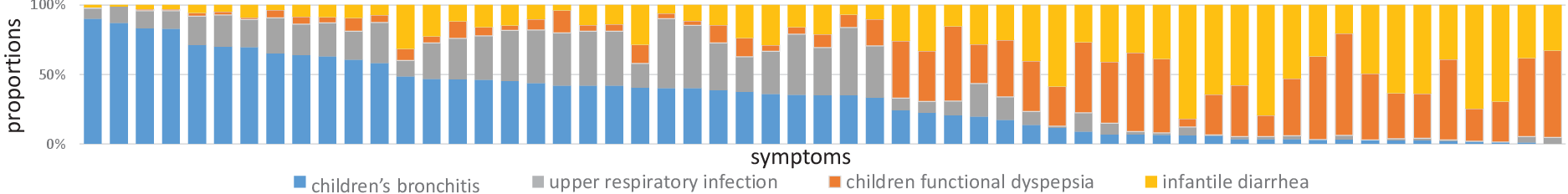}
\caption{Disease distribution over symptoms in MZ-4 (see section \ref{sec-rs-dataset}). X-axis stands for symptoms and y-axis is the proportion. Each bar describes the disease distribution given a symptom. }
\label{figure:symptomdistribution}
\end{center}

\end{figure*}

In the actual diagnosis scenario, we find that the relationship between diseases and symptoms can help us classify the disease. In Figure~\ref{figure:symptomdistribution}, we present the proportion of symptoms related to four different diseases, i.e., children bronchitis, upper respiratory infection, children functional dyspepsia and infantile diarrhea. It shows that a particular disease is usually related to a certain group of symptoms. In offline consultation, doctors also do the pre-examination and triage according to the different symptoms that the patient suffered, then doctors in different departments will make a more detailed diagnosis. This method significantly reduces the workload of individual doctors and enables them to be more specialized in a certain field. 

Hierarchical Reinforcement Learning (HRL)~\citep{ronald1998advance,richard1999}, in which multiple layers of policies are trained to perform decision making, conforms to the problem-solving logic for disease diagnosis in the real environment. HRL has been successfully applied to different scenarios, inter alia, course recommendation~\citep{zhang2019hierarchical}, visual dialogue~\citep{zhang2018multimodal}, relation extraction~\citep{feng2018relation} and composite tasks with slot constraints~\citep{lipton2018bbq}. In each step, the agent chooses either a one-step ``primitive" action or a ``multi-step" action (option). \citep{schatzmann2007agenda} presented a POMDP dialogue system for simulating user behavior, which can train and test a prototype system. Then, researchers showed that hierarchical reinforcement learning dialogue agents are feasible and promising in large-scale systems ~\citep{cuayahuitl2010evaluation}.
In order to improve the generalization ability of HRL model, ~\citep{carlos2017snn} proposed a general framework that first learns useful skills (high-level policies) in an environment and then leverages the acquired skills for learning faster in downstream tasks. Then, \citep{budzianowski2017sub} applied HRL to multi-domain dialogue management, which showed the potential of HRL to facilitate policy optimization for more sophisticated multi-domain dialogue systems. \citep{lipton2018bbq} used HRL to efficiently learn the dialogue manager that operates at different temporal scales.
Up to now, HRL has been successfully applied to different tasks and reached promising results.~\citep{wang2018video,zhang2018multimodal,takanobu2018hierarchical,feng2018relation,guo2018long, zhang2019hierarchical, wan2020reasoning, duan2020hierarchical}. Most existing works decompose the corresponding task into two steps manually, where the first step is treated by the high-level policy while the second step is treated by the low-level policy. This motivates us to divide the online diagnosis tasks into different levels following the setting of departments in the hospital and design a hierarchical structure for symptom acquisition and disease diagnosis.

In this paper, we propose to build a dialogue system with a hierarchy of two levels for automatic disease diagnosis using HRL methods. The high-level policy consists of a model named master and the low-level policy consists of several workers and a disease classifier. The master is responsible for triggering a model at a low level. Each worker is responsible for inquiring about symptoms related to a certain group of diseases while the disease classifier is responsible for making the final diagnosis based on information collected by workers. The proposed framework imitates a group of doctors from different departments diagnosing a patient together. Among them, each worker acts like a doctor from a specific department, while the master acts like a committee that appoints doctors to interact with this patient. When the information collected from workers is sufficient, the master would activate a separate disease classifier to make the diagnosis. Models in the two levels are trained jointly for better disease diagnosis. We test our model in three large real-world datasets and a synthetic dataset. Experimental results demonstrate that the performance of our hierarchical framework outperforms other state-of-the-art approaches.

We summary our contribution as follows: 1) We propose a new RL-based method for automatic diagnosis. It simulates the real scene of clinical practice, and assigns patients to different workers through high-level policy, thereby reducing the action space and improving training efficiency. Also, the method can be compatible with different network models and training strategies. 2) We perform systematical evaluation to test the performance of our model on three public datasets from the real environment and a synthetic datasets. Overall experiment results show the advantage of our model compared to state-of-the-art baselines and further analysis confirms the effectiveness of each component in our framework. 3) We release a toolkit with the implementation of all existing baseline models and datasets. The toolkit can be used as a benchmark for dialogue-based disease diagnosis.

\section{Materials and Methods}
In this section, we introduce our hierarchical reinforcement learning framework for disease diagnosis. We start with the formulation of Markov Decision Process for automatic disease diagnosis, and then introduce the hierarchical setting.

\subsection{Markov Decision Process Setup for Disease Diagnosis}
As for RL-based models for automatic diagnosis, the action space of agent is $\mathcal{A}=\mathit{D}\cup \mathit{S}$, where $\mathit{D}$ is the set of all diseases and $\mathit{S}$ is the set of all symptoms that associated with these diseases. Given the state $s_t\in \mathcal{S}$ at turn $t$, the agent takes an action according to it's policy $a_t\sim\pi(a|s_t)$ and receives an immediate reward $r_t=R(s_t, a_t)$ from the environment (patient/user). If $a_t\in\mathit{S}$, the agent chooses a symptom to inquire the patient/user. Then the user responds to the agent with \emph{true/false/unknown}. If $a_t\in \mathit{D}$, the agent informs the user of the corresponding disease as the diagnosis result and the dialogue session will be terminated as the success/failure in terms of the correctness of the diagnosis.

\subsection{Hierarchical Policy Structure for Disease Diagnosis}

\begin{figure*}[t]
\centering
\subfigure[Master-worker framework (training)]{
\centering
\includegraphics[width=6cm]{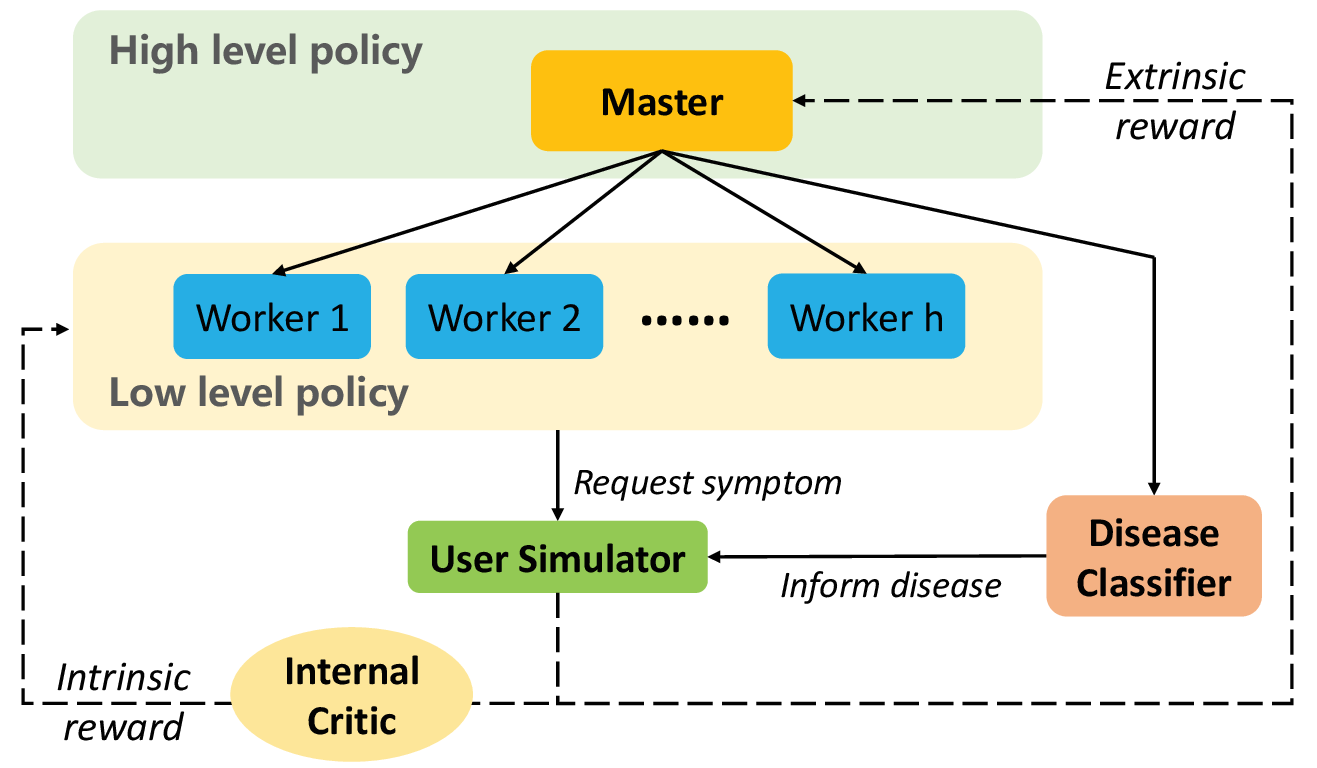}
}
\subfigure[Master-worker framework (inference)]{
\centering
\includegraphics[width=9.6cm]{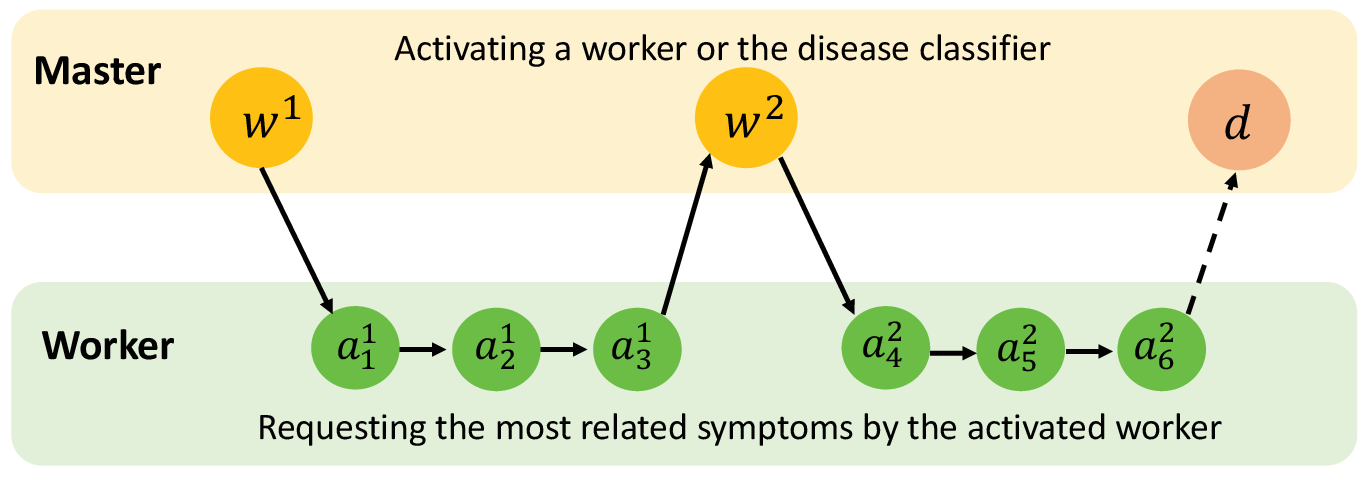}
}
\caption{The training and inference process of our model. (a) The framework of our hierarchical reinforcement learning model with two-layer policies. (b) Illustration of the diagnosis process of our model with interactions between components of two levels. $w^i$ is the action invoking worker $w^i$ and $d$ is the action invoking disease classifier. }
\label{diagnosisprocess}
\end{figure*}

To reduce the large action space, we extend the flat-RL structure to a hierarchical structure with two-layer policies for automatic diagnosis. Following the \emph{options} framework~\citep{richard1999}, our framework is illustrated as in Figure~\ref{diagnosisprocess}(a). There are four components in our framework: master, worker, disease classifier, and user simulator. At turn $t$, the state $s_t$ will be encoded as one-hot vectors that reflect the status of each symptom and number of turns for the master and worker network, and only symptom information will be encoded for the disease classifier. An illustration of the diagnosis process with interactions between models in two levels is presented in Figure~\ref{diagnosisprocess}(b). 


\subsubsection{Master for classification}
The master controls the higher policy of the agent. At each turn, the master can choose whether to activate the worker to collect symptom information or disease classifier to make a decision. Once the master activates a worker, this worker will interact with the user for $N$ turns until the subtask is terminated. 


The learning problem of the master can be formulated as a Semi-Markov Decision Process (SMDP), where the extrinsic rewards returned can be accumulated as the immediate rewards for the master~\citep{ghavamzadeh2005hierarchical}. That is to say, after taking an action $a_t^m$, the reward $r_t^m$ for the master can be defined as:
\begin{equation}
    \label{commulatedreward}
r^m_t=\left\{\begin{array}{cl}
\sum_{t'=1}^N \gamma^{t'} r_{t+t'}^{e},& if \quad a^m_t=w^i\\[2pt]
r_t^e, &if \quad a^m_t=d
\end{array}
\right.
\end{equation}
    
where $\gamma$ is the discounted factor and $w^i$, $d$ is the action to activate the worker $w^i$ and disease classifier. The objective of the master is to maximize the extrinsic reward, thus we can write the master's loss function as follows:
\begin{equation}
    \mathcal{L}(\theta_{m})=\mathbb{E}_{s,a^m,r^m,s'\sim\mathcal{B}^m}[(y-Q_m(s,a^m;\theta_{m}))^2]
\end{equation}

where  $y=r^m+\gamma^N\max_{{a^{m}}'}Q_m(s',{a^{m}}';\theta_{m}^{-})$, $\theta_{m}$, $\theta_{m}^{-}$ is the network parameter at current and previous iteration, $\mathcal{B}^m$ is the fixed-length buffer of samples for master.

\subsubsection{Worker \& Disease Classifier for interaction}

The worker controls the lower policy of the agent and interacts with the patient to collect information for a specific group of symptoms. Once the worker $w^i$ is invoked, it will take the corresponding state representation $s^i$ from the master and generate an action $a^i\in \mathcal{A}^w_i$.

After taking action $a^i\in\mathcal{A}^w_i $, the state representation will be updated and worker $w^i$ will receive an intrinsic reward $r_t^{i}$ from the user simulator. So the objective of workers is to maximize the expected cumulative discounted intrinsic rewards. The loss function of worker $w^i$ can be written in the following way:
\begin{equation}
    \mathcal{L}(\theta^i_{w})=\mathbb{E}_{s^i,a^i,r^{\mathtt{i}},{s^i}'\sim\mathcal{B}^w_i}[(y_i-Q_w^i(s^i,a^i;\theta^i_{w}))^2]
\end{equation}

Once the disease classifier is activated by the master, it will take the symptom information as input and output a vector $\bold{p}\in \mathbb{R}^{|\mathit{D}|}$, which represents the probability distribution over all diseases. The disease with the highest probability will be returned to the user as the diagnosis result. Also, the disease classifier will be jointly trained with the master and worker through supervised learning.

\subsubsection{User Simulator \& internal critic for reward generation}

Following~\citep{wei2018task} and~\citep{xu2019end}, we set up a user simulator to interact with the agent. At the beginning of each dialogue session, the user simulator samples a user goal from the training set randomly. Each piece of user goal contains two kinds of symptoms, namely explicit ones (obtained from the self-report) and implicit ones (obtained from the dialogues). Explicit symptoms will be directly provided to the agent as initial information at the beginning, and the agent needs to discover the implicit symptoms during the interaction with the patient. The simulator will initialize the dialogue session based on the explicit symptoms and interacts with the agent based on the implicit symptoms. 

In addition, the internal critic will generate a reward to the master and worker due to the action and the dialogue status. In the higher level, the dialogue session will be terminated as successful and get a positive reward if the agent makes the correct diagnosis, or failed if the informed disease is incorrect or the dialogue reaches the maximal turn $T$. In the lower level, a worker is terminated as successful when a correct symptom is requested by the agent and failed when the number of turns reaches the upper limit of subtask turn $T^{sub}$.

\subsection{Implementation details}
To group diseases with similar symptoms, we formulate the dataset as a disease-symptoms vector, which represents the number of times that a symptom occurs in each disease. Then we compare the similarities of the vectors on the training set and divide the diseases with higher similarity (above 0.5) into one group.

Also, to solve the problem of sparse action space and force the master to choose an efficient worker, we follow \citep{peng2018refuel} and use the reward shaping method to add auxiliary reward to the original extrinsic reward while keeping the optimal reinforcement learning policy unchanged.

In practice, the $\epsilon$ for the master and all the workers are all set to 0.1. For the master, the maximal dialogue turns $T$ is set to 10, it will receive an extrinsic reward of +20 if the master informs the right disease. Otherwise, it will receive an extrinsic reward of -100 if the dialogue turn reaches the maximal turn. In other states, the extrinsic reward is 0. Moreover, the sum of the extrinsic rewards (after reward shaping) over one subtask will be the reward for the master. The maximal dialogue turns $T^{sub}$ is set to 2 for each worker.

For the master and all the workers, the neural network of DQN is a three-layer network with two dropout layers and the size of the hidden layer is 512, learning rate for the DQN network is set to 0.0005. All parameters are set empirically and settings for the datasets are the same. For the disease classifier, the neural network is a two-layer network with a dropout layer.  Moreover, it's trained every epoch during the training process of the master.

During the training process, it will take about 5000 epochs for the model to reach convergence, which takes about 18 hours given an NVIDIA RTX 2080 Ti. For the best-performing model,  $\gamma$ is set to 0.95, discounted factor $\gamma_w$ is set to 0.99,  $\lambda$ in reward shaping is set to +1. 

\section{Experiment Implementation} \label{section:Experiments}
We evaluate all methods on three dialogue datasets (\textbf{MZ-4}~\citep{wei2018task}, \textbf{MZ-10} and \textbf{Dxy}~\citep{xu2019end}) collected in the real environment and one synthetic dataset \textbf{SymCat-SD-90}. We construct MZ-10 as an expansion on the basis of MZ-4 by including more diseases and samples of patients. Table~\ref{table:discriptionofdataset} shows the details of all datasets used in this paper.

Also, we evaluate the performance of different methods in terms of disease and symptoms. In terms of disease, we use the accuracy of disease judgment as an indicator (\textbf{Acc.}). In terms of symptoms, we use match rate of symptoms (\textbf{M.R.}) which is the ratio of the number of corrected recalled implicit symptoms to the total number of implicit symptoms. At the same time, we report the average number of turns (\textbf{Avg. T}) conducted for dialogue sessions for reference. 

\subsection{Real-world Dataset}
\label{sec-rs-dataset}

\paragraph{MZ-4} ~ {This is the first dataset collected from a real environment for the evaluation of task-oriented dialogue system~\citep{wei2018task}. It includes 4 diseases, 230 symptoms, and 1,733 user goals. Each user record consists of the self-report provided by the user and conversation text between the patient and a doctor. Symptoms extracted from self-report are treated as explicit symptoms and the ones extracted from the conversation are implicit symptoms. The raw data is collected from the pediatric department on a Chinese online healthcare community\footnote{http://muzhi.baidu.com}, and annotators will follow the BIO(begin-in-out) schema for symptom identification. After that, experts manually link each symptom expression to a concept on SNOMED CT\footnote{https://www.snomed.org/snomed-ct}. }

\paragraph{Dxy} ~ {A Dialogue Medical dataset ~\citep{xu2019end} contains data from a Chinese online healthcare
website\footnote{https://dxy.com/}. They annotate five types of diseases, including allergic rhinitis, upper respiratory infection, pneumonia, children hand-foot-mouth disease, and pediatric diarrhea.
Also, they extract the symptoms and normalize them into 41 symptoms.
This dataset contains 527 user goals, including 423 for training and 104 for testing.}

\paragraph{MZ-10} ~ { It is expanded from MZ-4 to include 10 diseases, consisting of typical diseases of the digestive system, respiratory system, and endocrine system. Following~\citep{wei2018task}, we collect medical consultation records for 10 pediatric diseases. Then we annotate the samples to form the dataset. Based on the BIO schema, we tagged each symptom with an extra label: Positive, Negative, or Not Sure. Besides, we link all the symptoms to the most relevant concept on SNOMED-CT for normalization. For labeling, we developed a web-based tool and recruited undergraduates and postgraduates in medical school to annotate the corpus. All the annotators are people who are willing to participate and are over the age of 18. Each dialogue is annotated twice and inconsistent parts are further finalized by a third annotator. The kappa coefficient of symptom labels is 92.71\%, which represents a high consistency between the two annotations.}


\subsection{Synthetic Dataset}
The number of diseases and user goals in the real-world dataset is still limited. To show the effectiveness of the HRL method, we build a synthetic dataset (SD) following~\citep{kao2018context} for further analysis of the method, named \textbf{SymCat-SD-90}. It is constructed based on a symptom-disease database called SymCat\footnote{www.symcat.com}. There are 801 diseases in the database and we classify them into 21 departments (groups) according to the International Classification of Diseases (ICD-10-CM)\footnote{https://www.cdc.gov/nchs/icd/}. We choose 9 representative departments from the database, each department contains the top 10 diseases according to the occurrence rate in the Centers for Disease Control and Prevention (CDC) database.

In the SymCat database, each disease is linked with a set of symptoms, where each symptom has a probability indicating how likely the symptom is identified for the disease. Given a disease and its related symptoms, the generation of a user goal follows two steps. First, for each related symptom, we sample the symptom-based on the probability. Second, a symptom is chosen randomly to be the explicit one (same as symptoms extracted from self-report in RD) and the rest of the true symptoms are treated as implicit ones. 
\begin{table*}[t]
\caption{Overview of Datasets. We report the number of user goal, the number of diseases, the average number of implicit symptoms in each sample and the total number of symptoms in the dataset. } 
\label{table:discriptionofdataset}
\begin{center}
\begin{tabular}{lcccc} \hline
\multicolumn{1}{c}{\bf Name}  &  \multicolumn{1}{c}{\bf$\sharp$ of user goal} &  \multicolumn{1}{c}{\bf$\sharp$ of diseases} &  \multicolumn{1}{c}{\bf avg. $\sharp$ of im. sym.} & \multicolumn{1}{c}{\bf $\sharp$ of sym.}\\ 
\hline
MZ-4   & 1,733  & 4  & 5.46  & 230\\
MZ-10  & 4,116  & 10 & 6.60  & 331\\
Dxy    & 527   & 5  & 1.67  & 41\\
SymCat-SD-90 & 30,000& 90 & 2.60  & 266\\ \hline
\end{tabular}
\end{center}
\end{table*}
\subsection{Models for Comparison} 

We compare our model with some state-of-the-art baselines for disease diagnosis.

\paragraph{Flat-DQN~\citep{wei2018task}} {This is the first work that treats the dialogue-based disease diagnosis as an MDP problem and employ an one-layer policy structure based on DQN to choose actions in each dialogue turn.} 

\paragraph{REFUEL~\citep{peng2018refuel}} {This work proposes two tricks to improve the performance of flat-DQN, namely reward shaping and feature rebuilding. Reward shaping aims to encourage the agent to discover positive symptoms more quickly by increasing the reward assigned to a correct symptom enquiry and penalizing incorrect ones. Feature rebuilding is introduced as an auxiliary component in the training process that aims to re-construct ground-truth symptom given the current information.} 

\paragraph{KR-DS~\citep{xu2019end}}~{This model proposes to use external knowledge to further improve the diagnosis performance of DQN. The overall framework contains three components, namely DQN-based policy network, refinement module based on co-occurrence relationship between diseases and symptoms, and a conditional probability matrix based on knowledge graph. The final decision is made integrating output from the three components to choose actions in each diaglogue turn. }

\paragraph{GAMP~\citep{xia2020generative}}~{This model integrates the Generative Adversarial Network (GAN) with the reinforcement learning model. The DQN-based policy network is employed the generator to choose the action and a discriminator is trained to determine how good the action is with a discriminative reward. Moreover, an independent disease classifier is used to evaluate the contribution of the generated symptom (i.e., action) with a mutual information reward. Both rewards are combined as the final reward to update the policy network.}

\paragraph{HRL-pretrained~\citep{kao2018context}}~{This model utilizes a hierarchical policy structure of two levels. There are two major differences between their model and ours. Firstly, the master in the higher level and workers in the lower level are trained separately in their setup while we jointly train the master and workers to enforce the interaction between the two components. Secondly, the diagnosis of the disease is handed over to workers in their setup while we introduce an additional disease discriminator that helps to allow workers to focus on symptoms. To some extent, this model can be treated as a pipeline training version of our model.} 

In the supplementary material, we use a table to show the main differences between models. All these above mentioned methods haven't disclosed their codes. For a fair comparison and encourage the follow-up researchers, we reproduce the their algorithms and release a toolkit\footnote{https://github.com/FudanDISC/DISCOpen-MedBox-DialoDiagnosis}. It can be used as the benchmark for dialogue-based disease diagnosis.

\section{Results and Discussions}
We report the performance of different approaches on real-world datasets and the performance on the synthetic dataset. Then, we perform the ablation study to evaluate the effectiveness of different components in the model. After that we present two analysis results to reveal the stability of the disease classifier and the efficiency of workers.

\subsection{Performance on Real-world Datasets}
In this section, we test all methods on three real-world datasets and compare the effectiveness of different models in terms of accuracy, average turns, and match rate.
Table~\ref{table:resultsOfalldataset} shows the overall results. We have following findings.\\
- \ \textbf{Flat-DQN} has limited ability to extract symptom information. Model performance drops significantly as the number of symptoms rises.\\
- \ \textbf{REFUEL} performs better than \textbf{Flat-DQN} in symptom extraction and also has a slight improvement in disease diagnosis. This proves the effectiveness of reward shaping and feature rebuilding.\\
- \ Compared with other methods, \textbf{KR-DS} maintains a higher match rate and accuracy on the three datasets. This indicates the effectiveness of introducing external knowledge to the RL-based agent.\\
- \ The symptom extraction ability of \textbf{GAMP} declines significantly with the increase in the number of symptoms, which also limits its performance.\\
- \ Our proposed model \textbf{HRL} generates the best performance in symptom extraction and disease classification among all the models. This confirms the effectiveness of our proposed hierarchical model.

\begin{table*}[t]
 \caption{Overall performance on real-world datasets. We conduct each experiment 5 times, and the reported number is the average.  $-$ denotes missing numbers, and Acc., M.R, Avg. T are the abbreviations of \emph{Accuracy}, \emph{Match Rate} and \emph{Average Turns} respectively. To make the results comparable, we keep all settings except the agent policy the same. Numbers in \textbf{Bold} are the best in each column.}
 \label{table:resultsOfalldataset}
 \begin{center}
\begin{tabular}{l|ccc|ccc|ccc}
\hline
\multicolumn{1}{l|}{} &   & \multicolumn{1}{c}{Dxy}  &  & \multicolumn{1}{l}{}  & \multicolumn{1}{c}{MZ-4} &           &   & \multicolumn{1}{c}{MZ-10} &  
\\ \hline
Model & Acc.  & M.R. & Avg. T & Acc. & M.R. & Avg. T & \multicolumn{1}{c}{Acc.} & \multicolumn{1}{c}{M.R.} & \multicolumn{1}{c}{Avg. T}  \\ \hline
Flat-DQN & 0.731 & 0.110 & 1.96 & 0.681 & 0.062 & 1.27 & 0.408 & 0.047   & 9.75 \\
REFUEL & 0.721  & 0.186  & 3.11  & 0.716 & 0.215 & 5.01 & 0.505 & 0.262 & 5.50  \\
KR-DS  & 0.740  & 0.399 & 5.65 & 0.678 & 0.177 & 4.61 & 0.485 & 0.279 & 5.95 \\
GAMP   & 0.731  & 0.268  & 2.84  & 0.644  & 0.107 & 2.93 & 0.500 & 0.067 & 1.78 \\

\hline

HRL (w/o grouped) & 0.731 & 0.297 & 6.61 & 0.689 & 0.004 & 2.25 & 0.540 & 0.114 & 4.59 \\
HRL (w/o discriminator) & $-$ & \textbf{0.512} & 8.42 & $-$ & \textbf{0.233} & 5.71 & $-$ & \textbf{0.330} & 8.75\\
HRL (ours) & \textbf{0.779} & 0.424 & 8.61 & \textbf{0.735} & 0.229 & 5.08 & \textbf{0.556} & 0.295 & 6.99 \\  
\hline
Classifier Lower Bound & 0.682 & $-$ & $-$ & 0.671 & $-$ & $-$ & 0.532 & $-$ & $-$\\
Classifier Upper Bound & 0.846 & $-$ & $-$ & 0.755 & $-$ & $-$ & 0.612 & $-$ & $-$  \\\hline
\end{tabular}
\end{center}
\end{table*}


\subsection{Performance on Synthetic Dataset}


In order to further prove the effectiveness of our model, we conduct the experiment on synthetic dataset. Experiment results of different approaches on synthetic datasets are shown in Table~\ref{table:resultsOfStepOnSD}. Through the results, we can get following findings.\\
- \ It is difficult for models with a flat policy structure to generate a good results on this dataset. \textbf{Flat-DQN}, \textbf{KR-DS}, \textbf{REFUEL} and \textbf{GAMP} perform similarly to each other. Besides, Both \textbf{Flat-DQN} and \textbf{GAMP} tend to make the diagnosis with a very short length of dialogue process. 
- \ The performance of \textbf{HRL-pretrained} is much better than that of other methods designed with flat and monolithic structure. This indicates the effectiveness of the hierarchical framework.\\
- \ Our proposed model \textbf{HRL} generates the best performance among all the models. This confirms the stability of our model.



\subsection{Ablation Studies}

The result of ablation study is shown in Table~\ref{table:resultsOfalldataset}. We compare several versions of HRL models. We remove the master and merge diseases into a single group to form a model with flat policy structure (denoted as HRL (w/o grouped). Besides, we remove the separate disease discriminator and hand over the diagnosis action to workers (denoted as HRL (w/o discriminator)). Experiment results show that HRL (ours) performs better than HRL (w/o discriminator) and HRL (w/o grouped). This indicates the effectiveness of extra disease classifier and the hierarchical policy structure. It is noteworthy that after we remove the disease classifier, the match rate of HRL(w/o discriminator) performs better than HRL(ours) on all three real-world datasets. This maybe because the reward for correctly predicting disease is higher than the reward for correctly predicting symptoms. This makes the model tend to make a final diagnosis rather than a symptom inquiry when it is confident enough, causing the model to ignore some symptoms that have less impact on the result. Therefore, the match rate of HRL will be smaller than the version without the disease discriminator.

\subsection{Stability of Disease Classifier}



\begin{figure}[h]
\centering
\centering
\includegraphics[width=6.0cm]{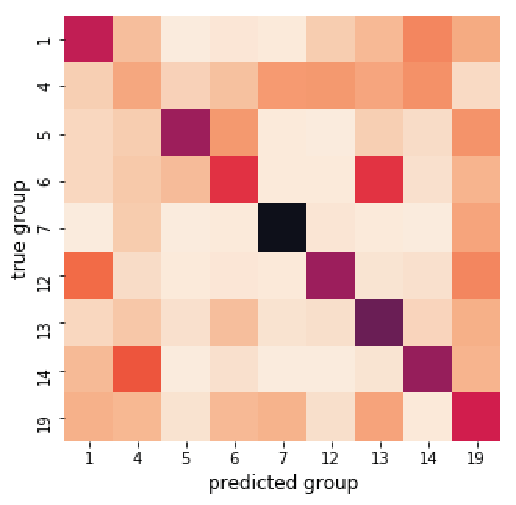}

\caption{The error analysis for the disease classifier. The square with true group i and predicted group j means a disease in group i is misclassified into group j, the darker the color, the higher the value. }
\label{groupanalysis}
\end{figure}

To have a deeper analysis of the user goals which have been informed of the wrong disease by the agent, we collect all the wrong informed user goals and present the error matrix in Figure \ref{groupanalysis}. It shows the disease prediction result for all the 9 groups. We can see the color of the diagonal square is darker than the others, which means the wrongly predicted disease and the correct disease are in the same group. This is reasonable because diseases in the same groups usually share similar symptoms and are therefore difficult to be distinguished. On the other hand, it also proves that even if the model cannot make correct predictions, it can still assist in real consultations. From ICD-10-CM, Group 7 (Diseases of the eye and adnexa) is prone to misjudgment within the group. For misjudgments between groups, the most likely ones are the Group 4 (Endocrine, nutritional and metabolic diseases) and Group 14 (Diseases of the genitourinary system), Group 6 (Diseases of the nervous system), and the Group 13 (Diseases of the musculoskeletal system and connective tissue).

It should be pointed out that although we say ``the disease classifier can assist in real consultations", this does not mean that the treatment for these diseases is the same. Diseases with similar symptoms may have completely different treatment options. In practice, we still need professional physicians to make the final decision. However, we believe that when the model can include more information (such as medical examinations, past medical history, etc.), more accurate judgments can be made.

\begin{table}[t]
 \caption{Overall performance on SymCat-SD-90 dataset. We conduct all experiments 5 times, and the reported number is the average. $-$ denotes missing numbers, and Acc., M.R, Avg. T are the abbreviations of \emph{Accuracy}, \emph{Match Rate} and \emph{Average Turns} respectively. Numbers in \textbf{Bold} are the best in each column. }
 \label{table:resultsOfStepOnSD}
  \begin{center}
\begin{tabular}{lccc} \hline
 \multicolumn{1}{l}{Model}    & \multicolumn{1}{c}{Acc.} & \multicolumn{1}{c}{M.R.} & \multicolumn{1}{c}{Avg. T}  \\\hline

 Flat-DQN & 0.343 & 0.023 & 1.23  \\
  KR-DS & 0.357 & 0.388 & 6.24 \\
  REFUEL & 0.347 & 0.161 & 4.56 \\
GAMP & 0.267 & 0.077 & 1.36 \\
\hline
 HRL-pretrained & 0.452 & $-$ & 3.42  \\
 Ours & \textbf{0.504} & \textbf{0.495} & 6.48  \\ 
\hline
Classifier Lower Bound & 0.308 & $-$ & $-$ \\
Classifier Upper Bound & 0.781 & $-$ & $-$\\\hline
\end{tabular}
\end{center}
\end{table}

\subsection{Efficiency of Different Workers}
\begin{table}[t]
\renewcommand\arraystretch{1.25}
\begin{center}
\small
\centering
\caption{The performance of different workers in SymCat-SD-90 dataset. }
\label{descriptionofworker}
\begin{tabular}{cccccc} \hline
Group id & Success rate & Ave intrinsic reward & Match rate & Activation times  \\ \hline

1 & 48.6\%& 0.031& 16.74\% &0.615  \\
4& 54.6\%&-0.150 & 5.02\%& 0.375 \\
5& 38.8\%&-0.013 & 7.96\%  &3.252 \\
6& 48.0\%&-0.036 & 9.58\% &0.942 \\
7& 48.3\%&0.057 & 18.57\% & 1.280 \\
12&43.0\%& 0.021 & 11.26\% & 0.666 \\
13& 52.4\%&-0.138 & 7.18\% & 0.823 \\
14& 72.2\%&-0.111 & 3.77\% & 0.614 \\
19& 47.4\%&0.031& 22.72\% & 1.124 \\\hline
Average & 50.3\% &-0.041& 10.49\% &1.077   \\
\hline
\end{tabular}
\end{center}
\end{table}
While proving the effect of the disease classifier, we also hope that the workers in each group can also play a positive role during the judgment. We evaluate the performance of workers in terms of success rate, average intrinsic rewards, and match rate. The results can be seen in Table~\ref{descriptionofworker}. We found most of the workers can successfully exit by querying the symptoms that which patient suffered. It proves that workers in different groups can learn the symptom characteristics of the group, and use the knowledge to guide the consultation process.

\section{Conclusion}
In this work, we formulate the problem of disease diagnosis as a hierarchical policy learning problem, where symptom acquisition and disease diagnosis are assigned to different kinds of workers at the lower level of the hierarchy. The experimental results on all datasets demonstrate that our hierarchical model outperforms other RL-based models in both disease accuracy and symptom recall. Since the input only contains symptom information, and the reinforcement learning model is sometimes not stable enough to have a statistically unbiased estimate of future expectations, the model cannot make a completely accurate diagnosis. But for now, we still believe that hierarchical architecture is the most reasonable structure in this field. At the same time, this method can also be integrated with other methods, such as the knowledge graph method, or adopting more advanced reinforcement learning techniques for master and worker. In the future, we would like to explore the dense representation of symptoms and diseases to improve the ability of generalization for automatic diagnosis. 

\section{Funding}
This work is partially supported by Natural Science Foundation of China (No.71991471, No.6217020551), Science and Technology Commission of Shanghai Municipality Grant (No.20dz1200600, 21QA1400600) and Zhejiang Lab (No. 2019KD0AD01).

\emph{Conflict of Interest: } The authors declare no competing interests.

\bibliographystyle{natbib}
\bibliography{document}

\end{document}